%% file: main.tex
\documentclass[runningheads]{llncs}

\usepackage{eccv}

\usepackage{eccvabbrv}

\usepackage{graphicx}
\usepackage{booktabs}

\usepackage[accsupp]{axessibility}  

\usepackage[pagebackref,breaklinks,colorlinks,citecolor=eccvblue]{hyperref}
\usepackage{enumitem, multirow, booktabs}
\usepackage{bm}
\usepackage{float}
\usepackage{wrapfig}
\usepackage{pifont}

\usepackage{orcidlink}

\newcommand{\mypar}[1]{\vspace{-2mm}\paragraph{{\bf #1}}}
\newcommand{\mysubsec}[1]{\vspace{-1.5mm}\subsection{#1}}

\begin{document}

\title{Self-Supervised Audio-Visual \\ Soundscape Stylization} 

\titlerunning{Self-Supervised Audio-Visual Soundscape Stylization}

\author{Tingle Li\inst{1} \and Renhao Wang\inst{1} \and Po-Yao Huang\inst{2} \and \\ Andrew Owens\inst{3}\and Gopala Anumanchipalli\inst{1}}

\authorrunning{T. Li et al.}


\institute{$^{1}$University of California, Berkeley
$^{2}$FAIR, Meta
$^{3}$University of Michigan \\
\url{https://tinglok.netlify.app/files/avsoundscape}
}

\maketitle

\input{sec0_abs}
\input{sec1_intro}
\input{sec2_related}
\input{sec3_method}
\input{sec4_exp}
\input{sec5_discussion}

%
%
\clearpage
\bibliographystyle{splncs04}
\bibliography{main}

\clearpage
\input{sec6_appendix}
\end{document}

%% file: sec0_abs.tex
\vspace{-2.5mm}
\begin{abstract}
Speech sounds convey a great deal of information about the scenes, resulting in a variety of effects ranging from reverberation to additional ambient sounds. In this paper, we manipulate input speech to sound as though it was recorded within a different scene, given an audio-visual conditional example recorded from that scene. Our model learns through self-supervision, taking advantage of the fact that natural video contains recurring sound events and textures. We extract an audio clip from a video and apply speech enhancement. We then train a latent diffusion model to recover the original speech, using another audio-visual clip taken from elsewhere in the video as a conditional hint. Through this process, the model learns to transfer the conditional example's sound properties to the input speech. We show that our model can be successfully trained using unlabeled, in-the-wild videos, and that an additional visual signal can improve its sound prediction abilities.
\end{abstract}

\vspace{-5mm}
\begin{figure}
    \includegraphics[width=\textwidth]{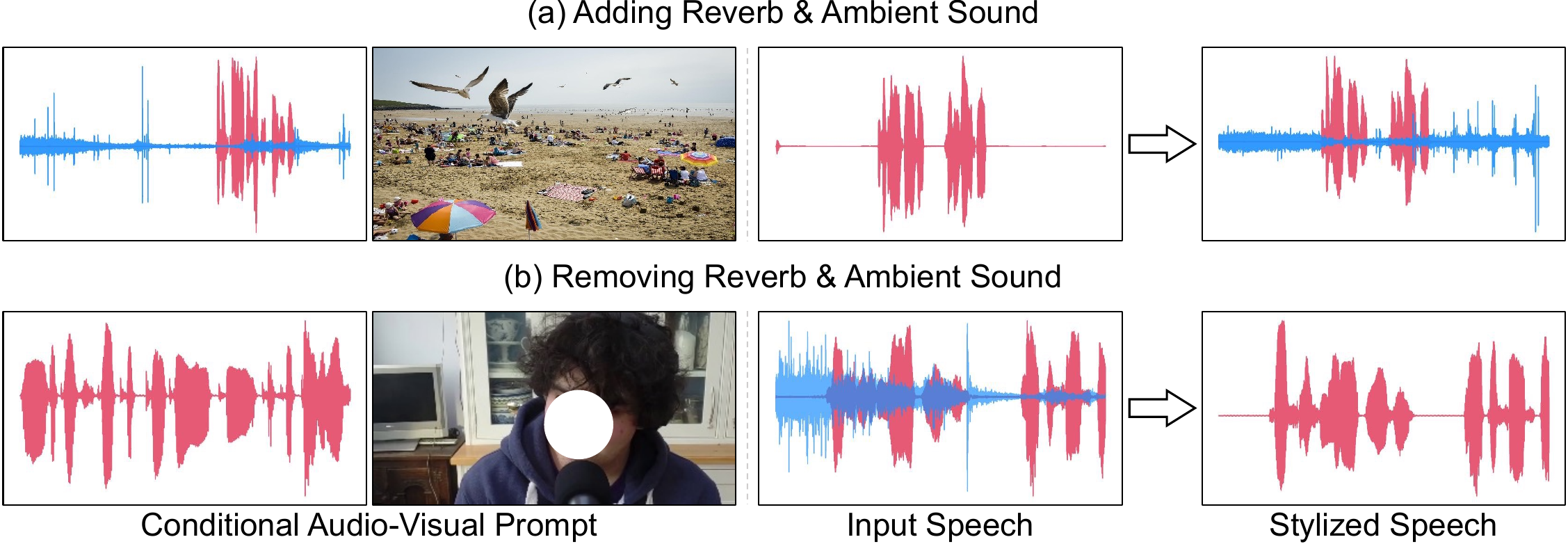}
    \caption{\small
    {\bf Audio-visual soundscape stylization.} 
    We learn through self-supervision to manipulate input speech (middle) such that it sounds as though it were recorded within a given scene (left). Our approach captures both acoustic properties, such as reverberation, as well as the ambient sounds, such as crashing waves (top). 
    To help convey the results of the stylization, we have used source separation to visualize the speech waveform (shown in \textcolor[rgb]{0.8627450980392157, 0.0784313725490196, 0.23529411764705882}{red}) separately from background sound (shown in \textcolor[rgb]{0.11764705882352941, 0.5647058823529412, 1.0}{blue}). 
    }
    \vspace{-6mm}
    \label{fig:teaser}
\end{figure}

%% file: sec1_intro.tex
\vspace{-2mm}
\section{Introduction}
\label{sec:intro}
Speech conveys a tremendous amount about the scene that it was produced in, from the material properties of its surfaces to its ambient sounds~\cite{pijanowski2011soundscape, grinfeder2022we}. A major goal of the audio and audio-visual generation communities has been to accurately resynthesize speech to sound as though it were recorded in a different scene~\cite{valimaki2016more, singh2021image2reverb, chen2022visual}, thereby capturing these subtle details --- a task that has a number of applications, from movie dubbing to virtual reality.

Existing formulations of this problem have largely focused on reproducing room acoustic properties, such as adding reverb by manipulating the room impulse response~\cite{chen2022visual, somayazulu2024self}.  However, these approaches do not model many of the other ways that a scene can affect a recorded sound. For example, when we walk on the beach (Figure~\ref{fig:teaser}), we may experience the whispers of the wind, the cries of seagulls, and the crash of waves --- a distinctive ambient sound texture~\cite{mcdermott2011sound} that would be conveyed in any sound recording taken within the scene. Since these aspects of a soundscape are not modeled by existing resynthesis techniques, making a sound fully reflect a scene requires additional postprocessing, such as ``mixing in''  background noises that fit the scene. This process, often reliant on descriptive language, can be time-consuming and constrained in its ability to convey subtle auditory properties. Moreover, these methods have largely required simulated (or labeled) training data, and are not designed to learn from abundantly available ``in the wild'' audio-visual data, limiting their scalability.

We propose the {\em audio-visual soundscape stylization} problem. Given a visual or audio-visual example from a scene and a clean input speech, our goal is to manipulate the input speech such that it could have occurred within the scene, reproducing both the acoustic properties of a scene and the ambient sounds within it. 
Our method is based on {\em conditional speech de-enhancement}:
we randomly sample two nearby audio-visual clips from a video, and remove scene-specific attributes from one of them performing speech enhancement. We then train a model, based on latent diffusion~\cite{rombach2022high}, to reverse this speech enhancement process, using the other audio-visual clip as a conditional ``hint.''  In order to perform this task, the model needs to infer the acoustic and ambient properties of the scene, and to successfully transfer them to an input speech. At test time, we give the model a conditional example from the scene whose properties we would like to transfer.

Our model is simple and can be trained entirely using in-the-wild egocentric videos. We show through quantitative evaluations and perceptual studies that our method learns to successfully stylize sounds in a number of challenging in-the-wild scenarios, transferring both the acoustic properties and ambient sounds to input speech.   
As part of these experiments, we find that our model can successfully transfer sound from visual conditioning, and that visual signals improve our model's ability to stylize audio. We also find that we can outperform existing work on the previously proposed problem of styling sound using room acoustic properties from images~\cite{chen2022visual} while going beyond this work in also transferring ambient sound. Finally, we find that ``prompting'' our model with specific conditional examples can achieve a desired style, such as approximately converting between near- and far-field speech, and that our model can successfully restyle a variety of non-speech input sounds.

%% file: sec2_related.tex
\section{Related Work}
\label{sec: related_work}

\mypar{Stylization in image and audio.}
The concept of image stylization was pioneered by Hertzmann et al. \cite{hertzmann2023image}, which restyled input images based on a single user-provided example.
Various types have been explored for image stylization, including image \cite{huang2017arbitrary, isola2017image, zhu2017unpaired}, text \cite{dong2017semantic, bau2021paint, brooks2023instructpix2pix}, sound \cite{li2022learning, lee2022sound}, and touch \cite{yang2022touch, yang2023generating}.
Recent work has addressed a variety of audio stylization tasks, such as voice conversion (via feature disentanglement \cite{Ulyanov2016audio, verma2018neural} or adversarial learning \cite{kaneko2018cyclegan, li2020cvc}), 
music timbre transfer \cite{huang2018timbretron},  text-driven audio editing \cite{wang2023audit}, visual acoustic matching \cite{chen2022visual, somayazulu2024self}, and audio effects stylization \cite{steinmetz2022style}. In particular, Chen et al. \cite{chen2022visual} proposed to manipulate the room impulse response based on the surrounding images using a generative adversarial network (GAN). Recently, Somayazulu et al. \cite{somayazulu2024self} used an additional GAN to further denoise the target audio to get the paired data for training. In contrast to these works, our method differs by: (\romannumeral1) Generating ambient sounds beyond mere room impulse response; (\romannumeral2) Using a more expressive diffusion model rather than GAN; (\romannumeral3) Learning from ``in-the-wild'' internet videos instead of curated indoor videos.

\mypar{Sound generation from visual and textual inputs.}
Generating sound from visual and textual inputs has recently attracted much research attention. 
For visual-based methods, 
researchers have explored the generation of sound effects, music, speech, and ambient sound from visual cues such as impacts \cite{owens2016visually}, musical instrument playing \cite{koepke2020sight}, dancing \cite{gan2020foley, su2021does},  lip movements \cite{ephrat2017vid2speech, prajwal2020learning, hu2021neural}, and open-domain images \cite{zhou2018visual, iashin2021taming, sheffer2023hear, luo2023diff}. 
In contrast to these methods, which focus on generating a specific type of sound from visual input, our method is centered on stylizing soundscapes to fit their environmental context, guided by conditional audio-visual examples.
For text-based methods, Yang et al. \cite{yang2023diffsound} introduced a discrete diffusion model for generating ambient sound from text descriptions. Kreuk et al. \cite{kreuk2023audiogen} used VQGAN \cite{esser2021taming} for sound generation. 
Recently, latent diffusion models~\cite{liu2023audioldm, huang2023make} have significantly improved generation quality. 
All these methods necessitate text annotations to establish text-audio pairs, whether at the training \cite{kreuk2023audiogen} or representation \cite{liu2023audioldm, huang2023make} level, which can be labor-intensive and limited in expressiveness with plain text descriptions. 
In contrast, we make an input speech match a given soundscape that is specified by a given audio-visual example, without the need for annotations or text. This allows users to select the (often difficult-to-articulate) auditory properties ``by example'' rather than through language. Our approach thus provides a complementary learning signal to text-based methods.

\mypar{Audio-visual learning.}
The natural correlation between audio and visual frames in videos has facilitated extensive audio-visual research, including representation learning~\cite{owens2016ambient, arandjelovic2017look, korbar2018cooperative, owens2018audio, patrick2021space, morgado2021audio, huang2023mavil}, source separation \cite{zhao2018sound, zhao2019sound, ephrat2018looking, gao2018learning, li2020atss}, audio source localization \cite{chen2021localizing, harwath2018jointly, chen2023sound}, audio spatialization \cite{gao20192, morgado2018self, yang2020telling}, visual speech recognition \cite{afouras2018deep}, deepfake detection \cite{feng2023self}, and scene classification \cite{chen2020vggsound, gemmeke2017audio, du2023uni}. 
Inspired by this line of work, we aim to stylize input audio to match the soundscape of a single audio-visual conditional example.

%% file: sec3_method.tex
\section{Audio-Visual Soundscape Stylization}
\label{sec:method}
Our goal is to manipulate an input sound such that it could plausibly have been recorded within another scene, given a conditional audio-visual example from that scene. We learn a function $\mathcal{F}_\theta(\bm{a}_e, \bm{a}_c, \bm{i}_c)$ parameterized by $\theta$ that stylizes an input sound $\bm{a}_e$ given the reference audio $\bm{a}_c$ and its corresponding image $\bm{i}_c$. We show that $\mathcal{F}_\theta$ can be learned solely from unlabeled audio-visual data. 

\subsection{Self-Supervised Soundscape Stylization}
\label{subsec:pretext}

We propose a self-supervised task that trains a model to stylize input sounds, using audio-only, visual-only, or audio-visual conditioning. 
\mypar{Learning by audio-visual speech de-enhancement.} 

\begin{wrapfigure}[18]{r}{0.6\textwidth}
    \centering
    \vspace{-8mm}
    \includegraphics[width=\linewidth]{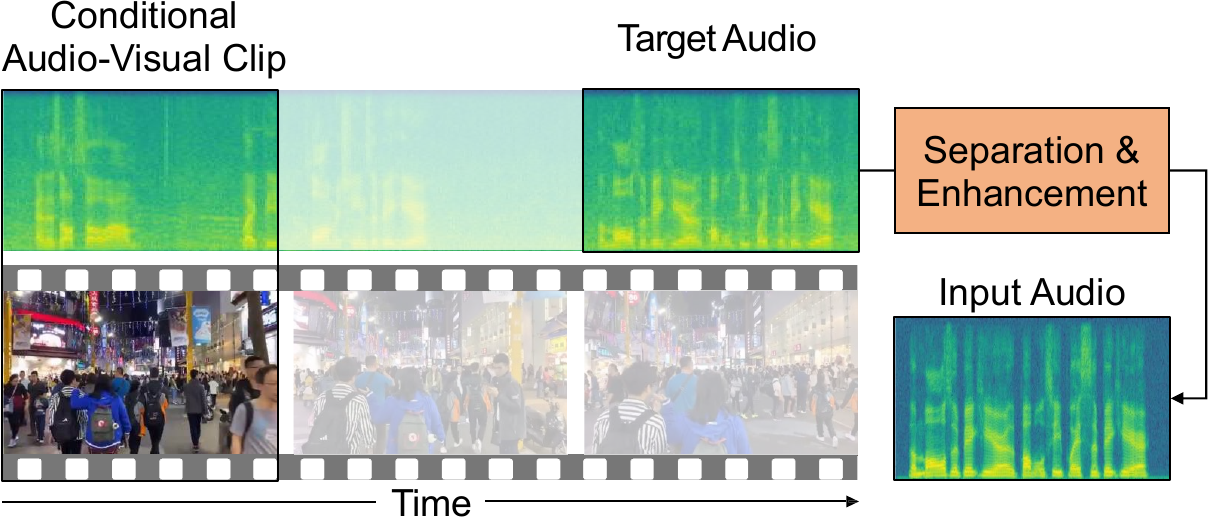}
    \caption{{\bf Soundscape stylization by conditional speech de-enhancement}. We randomly select two disjoint clips from a video, designating one as a {\em conditional} example and the other as the {\em target}. We then separate and enhance the target audio. Our model's self-supervised pretext task is to remove this enhancement using the other conditional (audio, visual, or audio-visual) signal as a hint. At test time, we stylize an audio clip using a conditional example from the desired scene.}
    \label{fig:pretext}
\end{wrapfigure}
As a self-supervised pretext task~\cite{doersch2017multi}, we randomly select an audio clip from a training video, apply source separation and enhancement to it, and train a model to undo this operation (\ie, to recover the original sound) after conditioning on another audio-visual example taken from the same training video (Fig.~\ref{fig:pretext}).
We observe that the background noises and acoustic properties within a video tend to exhibit temporal coherence \cite{du2023conditional}, especially when sound events occur repeatedly. Moreover, similar sound events often share semantically similar visual appearances~\cite{owens2016ambient, huh2023epic}.
By providing the model with a conditional example from another time step in the video, the model is implicitly able to estimate the scene properties and transfer these to the input audio (\eg, the reverb and ambient background sounds). At test time, we will provide a clip taken from a {\em different} scene as conditioning, forcing the model to match the style of a desired scene. 

Specifically, we sample non-overlapping clips from a long video, centered at times $\tau$ and $\tau^{\prime}$. One of these clips serves as the conditional audio-visual example, denoted as $\bm{a}_c$ and $\bm{i}_c$, while the soundtrack of the other clip is designated as the target audio $\bm{a}_q$. We then apply both a source separation model~\cite{petermann2022cocktail} to isolate the foreground speech, and a speech enhancement and dereverberation model~\cite{adobe2023enhance} to further refine the separated speech quality. This process results in the generation of high-fidelity speech $\bm{a}_e = \mathcal{H}(\bm{a}_q)$, which sounds as if it were recorded in a soundproofed studio. Please refer to the Appendix \ref{app:enhance} for the analysis of different enhancement strategies. For preprocessing, we use an off-the-shelf voice activity detector \cite{silerovad2021} to ensure that each selected audio clip is likely to contain speech.

\mypar{Sound stylization model.}
After training on the pretext task of audio-visual speech de-enhancement, the resulting model is able to tailor its stylization according to the conditional examples, which aligns with the assumption that the conditional example is instructive for the input audio. At test time, we retain the flexibility to substitute the conditional example with a completely different audio-visual clip, enabling the potential for one-to-many stylization.

\begin{figure*}[t]
    \centering
    \includegraphics[width=\linewidth]{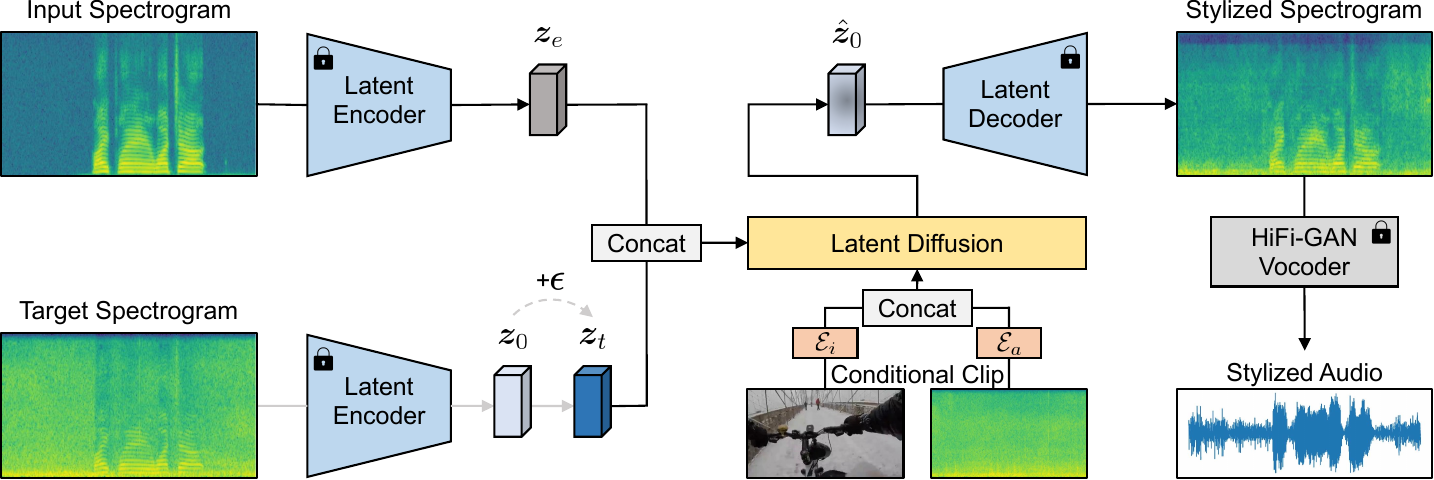}
    \caption{{\bf Model architecture}. Given input audio derived from an enhancement model, and the conditional audio-visual clip sampled from the same video, we aim to stylize the input to closely resemble the original signal. We encode both the input and target spectrograms to the latent space using a pre-trained latent encoder, and feed them into a latent diffusion model together with the conditional audio-visual embedding. The goal is to harmonize the encoded latent of the input spectrogram with the target one. Finally, we employ a pre-trained latent decoder followed by a pre-trained HiFi-GAN vocoder to reconstruct the waveform from the latent space. Note that the latent encoder for the target spectrogram is {\em not} used at test time.}
    \label{fig:model}
    \vspace{-4mm}
\end{figure*}

\mysubsec{Conditional Stylization Model}
\label{subsec:diffusion}
We describe our conditional soundscape stylization model $\mathcal{F}_\theta$ (Figure~\ref{fig:model}), which is designed to stylize input audio based on a conditional audio-visual pair and consists of three main components: \romannumeral1) compressing the input audio into a latent space; \romannumeral2) applying soundscape stylization using the conditional latent diffusion model; \romannumeral3) reconstructing the waveform from the latent space.

\mypar{Conditional latent diffusion model.}
We train a conditional latent diffusion model to stylize soundscapes based on conditional audio-visual examples. Building upon the denoising diffusion probabilistic model \cite{ho2020denoising} and the latent diffusion model \cite{rombach2022high}, our model breaks the generation process into $N$ conditional denoising steps, and improves the efficiency of diffusion models by operating in the latent space. Therefore, our soundscape stylization model $\mathcal{F}_\theta$ can be interpreted as an equally weighted sequence of denoising auto-encoders $\bm{\epsilon}_\theta$.

Specifically, it takes the encoded latent of the input audio and the conditional audio-visual example as a conditioning signal. To elaborate further, given the encoded latent of the original audio $\bm{z}_0 = {\rm Enc}(\bm{a}_q)$, the conditional audio-visual example $(\bm{a}_c, \bm{i}_c)$, a random denoising step $t$, and random noise $\bm{\epsilon} \sim \mathcal{N}(\mathbf{0}, \mathbf{I})$, our model first generates a noisy version $\bm{z}_t$ via a noise schedule \cite{song2020denoising}. We then define the training loss $\mathcal{L}_\theta$ by predicting the noise $\bm{\epsilon}$ added to the noisy latent, guided by the input audio $\bm{a}_e$ and the conditional audio-visual pair $(\bm{a}_c, \bm{i}_c)$. This can be achieved by minimizing the loss function as follows:
\vspace{-1.5mm}
\begin{equation}
\label{eq:loss_diffusion}
    \mathcal{L}_\theta = \mathbb{E}_{\bm{z}_0,\bm{a}_c, \bm{i}_c, \bm{\epsilon} \sim \mathcal{N}(\mathbf{0}, \mathbf{I}), t}\Vert\bm{\epsilon} - \bm{\epsilon}_\theta(\bm{z}_t, t, \bm{z}_{e},\bm{a}_c, \bm{i}_c)\Vert^{2}_{2}\text{ ,}
\end{equation}
where $\bm{z}_{e} = {\rm Enc}(\bm{a}_e)$ is the encoded latent of the input audio $\bm{a}_e$.

\mypar{Adding noise to the input.}
We propose to add Gaussian noise $\bm{n} \sim \mathcal{N}(\mathbf{0}, {\sigma}^2\mathbf{I})$ to the enhanced audio at both training and test time. 
This mixed audio is then employed as the input audio. The primary purpose of this is to mitigate the effect of ``audio nostalgia'' \cite{donahue2023singsong}. This conceals subtle remnants of the original sound that may persist in the enhanced audio, which could potentially help avoid leaking information in the pretext task. Specifically, when the model is trained on the enhanced audio, the generated soundscapes might exhibit a suspicious resemblance to the originals. Conversely, when clean speech is used as input, the output may largely replicate the input. We consider this addition of noise as a type of data augmentation.

\mypar{Compressing mel-spectrograms.}
We employ a ResNet-based variational auto-encoder (VAE) \cite{he2016deep, kingma2013auto} to compress the mel-spectrogram $\bm{a} \in \mathbb{R}^{T \times F}$ into a latent space $\bm{z} \in \mathbb{R}^{T/r \times F/r \times d}$, where $r$ denotes the compression level, $T/r$ and $F/r$ is a lower-resolution time-frequency bin, and $d$ represents the embedding size at each bin. The VAE is tasked with reconstructing sounds from a dataset, where the bottleneck can then serve as the encoded latent. For our experiments, we adopt a pre-trained VAE model from Liu et al. \cite{liu2023audioldm}.

\mypar{Conditional audio-visual representations.}
The conditional audio-visual example is represented using its latent vector. We employ separate audio and image encoders, denoted as $\mathcal{E}_a(\cdot)$ and $\mathcal{E}_i(\cdot)$, to extract audio embeddings $\mathcal{E}_a(\bm{a}_c) \in \mathbb{R}^L$ and image embeddings $\mathcal{E}_i(\bm{i}_c) \in \mathbb{R}^L$, where $L$ represents the embedding size. We use pre-trained encoders like CLIP \cite{radford2021learning} and CLAP \cite{elizalde2023clap} for image and audio representation respectively. Prior to fusion, we apply linear projections to the image and audio embeddings, followed by concatenating and feeding them into the diffusion model through cross-attention mechanism \cite{vaswani2017attention}.

\mypar{Classifier-free guidance.}
We use classifier-free guidance \cite{ho2022classifier} to balance the trade-off between the quality and diversity of generated samples. It involves jointly training the model for both conditional and unconditional denoising. During training, we randomly nullify the conditional model with a fixed probability of 10\%. At test time, a guidance scale ($\lambda \geq 1$) is utilized to adjust the score estimates, skewing them towards the conditional $\bm{\epsilon}_\theta(\bm{z}_t, t, \bm{z}_{e},\bm{a}_c, \bm{i}_c)$ and away from the unconditional $\bm{\epsilon}_\theta(\bm{z}_t, t, \bm{z}_{e}, \varnothing, \varnothing)$.

\vspace{-4mm}
\begin{equation}
\label{eq:cfg}
\begin{split}
    \tilde{\bm{\epsilon}}_\theta(\bm{z}_t, t, \bm{z}_{e},\bm{a}_c, \bm{i}_c) = & \ \lambda \cdot \bm{\epsilon}_\theta(\bm{z}_t, t, \bm{z}_{e},\bm{a}_c, \bm{i}_c) \\
    & + (1-\lambda) \cdot \bm{\epsilon}_\theta(\bm{z}_t, t, \bm{z}_{e}, \varnothing, \varnothing)
\end{split}
\end{equation}

We find this guidance enhances the output quality and relevance of stylized samples.

\mypar{Recovering the waveform.}
Following the estimation of noise $\tilde{\bm{\epsilon}}_\theta$ from the diffusion model, we retrieve the encoded latent of the stylized mel-spectrogram. This latent is then fed into the VAE decoder to reconstruct the stylized mel-spectrogram. Finally, a pre-trained HiFi-GAN vocoder \cite{kong2020hifi} is employed to reconstruct the waveform.

%% file: sec4_exp.tex
\section{Experiments}

\subsection{Experimental Setup}
\label{subsec:exp_setup}

We evaluate our model's ability to restyle sounds to match new environments.

\mypar{Dataset.}
Our goal is to train and evaluate the model on minimally curated ``in-the-wild'' videos. To do this, we train and evaluate on two different datasets:  {\em CityWalk} and {\em Acoustic-AVSpeech}. 

\begin{itemize}[topsep=0pt, noitemsep, leftmargin=*]
    \item {\bf {\em CityWalk} dataset}: We collect a {\em CityWalk} dataset which includes egocentric videos with diverse ambient sounds and acoustic properties, recorded in places like trains, buses, streets, beaches, shopping malls, \etc. Using the search terms ``City Walk + POV'' on YouTube, we gather 3,447 videos derived from indoor (28\%) and outdoor (72\%) scenes. From this collection, we choose a subset of 235 videos for training and testing, with lengths varying between 5 to 225 minutes, totaling 158 hours. We ensure that these videos only contain naturally occurring sounds in the scenes, without any post-edited voice-overs or music. We also guarantee that the sources of training and testing videos do not overlap. Please see the Appendix \ref{app:dataset} for more dataset details.
    \item {\bf {\em Acoustic AVSpeech} dataset} \cite{chen2022visual}: The {\em Acoustic-AVSpeech} dataset is a subset of the AVSpeech dataset \cite{ephrat2018looking} that contains 3-10 seconds indoor clips of single speakers without interfering ambient sound. Since the visual content of these clips offers useful insights into the geometry and materials of the scenes, it can be utilized to estimate the acoustic properties (but not ambient sound). We use this dataset for a fair comparison with the existing baseline \cite{chen2022visual} (we are unable to compare with \cite{somayazulu2024self} as it is not open source).
\end{itemize}

\mypar{Model configurations.}
We use the VAE and HiFi-GAN vocoder from \cite{liu2023audioldm}, which are trained on the combination of AudioSet \cite{gemmeke2017audio}, AudioCaps \cite{kim2019audiocaps}, BBC Sound Effect \cite{bbc2023sfx} and Freesound \cite{fonseca2021fsd50k} datasets. The VAE is configured with a compression level $r$ of 4 and latent channels $d$ of 8. For extracting audio and image embedding, we have two options: \romannumeral1) use a from-scratch ResNet-18 encoder; \romannumeral2) utilize a fine-tuned CLAP audio encoder \cite{elizalde2023clap} derived from our audio-only model, alongside a fixed CLIP image encoder \cite{radford2021learning}. These encoders are integrated into the diffusion model through late fusion \cite{wang2020makes} and cross-attention \cite{vaswani2017attention}. The diffusion model is based on a U-Net backbone, consisting of four encoder and decoder blocks with downsampling and upsampling and a bottleneck in between. Multi-head attention with 64 head features and 8 heads per layer is applied in the last three encoder and first three decoder blocks. During the forward process, we employ $N = 1000$ steps and a linear noise schedule, ranging from $\beta_1 = 0.0015$ to $\beta_N = 0.0195$, to generate noise. Additionally, we leverage the DDIM sampling method \cite{song2020denoising} with 200 sampling steps. For classifier-free guidance, we set the guidance scale $\lambda$ to 4.5, as described in Equation \eqref{eq:cfg}.

\mypar{Training procedures.}
To enhance training efficiency, we divide all videos into 10-second video and audio clips. We apply a frame-level voice activity detector \cite{silerovad2021} to the resulting audio clips to detect speech onset. Subsequently, we randomly select two 2.56-second audio clips from the same source -- one for the target audio and the other for the conditional audio. The conditional image is chosen by randomly sampling one video frame within the scope of the selected conditional audio. Our model is trained using the AdamW optimizer \cite{loshchilov2017decoupled} with a learning rate of $10^{-4}$, $\beta_1 = 0.95$, $\beta_2 = 0.999$, $\epsilon = 10^{-6}$, and a weight decay of $10^{-3}$ over 200 epochs.

\mypar{Evaluation metrics.}
To assess the performance of our models, we use both objective and subjective metrics. Our objective metrics include the {\em Mean Square Error} (MSE), {\em RT60 Error} (RTE), {\em Mean Opinion Score Error} (MOSE), {\em Perceptual Evaluation of Speech Quality} (PESQ) \cite{rix2001perceptual}, {\em Fréchet Audio Distance} (FAD) \cite{kilgour2019frechet}, {\em Fréchet Distance} (FD) \cite{liu2023audioldm}, {\em Kullback-Leibler divergence} (KL), {\em Word Error Rate} (WER), {\em Inception Score} (IS) \cite{salimans2016improved}, and {\em Audio-Visual Correspondence} (AVC) \cite{arandjelovic2017look}. MSE evaluates how closely the stylized audio matches the ground truth in terms of magnitude spectrograms (if the ground truth is available), while RTE measures the MSE between RT60 estimates of generated and target speech, \ie, the differences in the reverb level. MOSE assesses the difference in speech quality between the generated audio and ground truth using MOSNet \cite{lo2019mosnet}. FD and FAD measure the similarity between real and generated audio using different classifiers (FAD employs VGGish \cite{hershey2017cnn} and FD uses PANNs \cite{kong2020panns}). KL quantifies the distributional similarity between real and generated audio, while PESQ and IS evaluate the quality and diversity of generated audio. WER measures the intelligibility of the generated audio using a pre-trained speech recognizer \cite{radford2023robust}. AVC assesses the correlation between audio and image, utilizing features extracted by either OpenL3 \cite{cramer2019look} or ImageBind (IB) \cite{girdhar2023imagebind}.

In addition, we conduct a subjective evaluation through Amazon Mechanical Turk. Human participants are asked to rate audio generated by various methods based on its similarity to the soundscapes in the given audio-visual example. This rate considers four criteria: {\em overall quality} (OVL), {\em relation to ambient sounds} (RAM), {\em relation to acoustic properties} (RAC), and {\em relation to visuals} (RVI), with scores ranging from 1 (low correlation) to 5 (high correlation). Please see the Appendix \ref{app:sub_eval} for more human evaluation details.

\begin{table}[t]
    \scriptsize
    \centering
    \setlength{\tabcolsep}{0.192mm}{
    \begin{tabular}{l c c c c c c c c c c}
    \toprule
    \multirow{2.5}{*}{\textbf{Method}} & \multirow{2.5}{*}{\textbf{MSE}\boldsymbol{\textsuperscript{*}}($\downarrow$)} & \multirow{2.5}{*}{\textbf{RTE}\boldsymbol{\textsuperscript{*}}($\downarrow$)} & \multirow{2.5}{*}{\textbf{PESQ}\boldsymbol{\textsuperscript{*}}($\uparrow$)} & \multirow{2.5}{*}{\textbf{FD ($\downarrow$)}} & \multirow{2.5}{*}{\textbf{FAD ($\downarrow$)}} & \multirow{2.5}{*}{\textbf{KL ($\downarrow$)}} & \multirow{2.5}{*}{\textbf{WER ($\downarrow$)}} & \multirow{2.5}{*}{\textbf{IS ($\uparrow$)}} & \multicolumn{2}{c}{\textbf{AVC ($\uparrow$)}} \\
    \cmidrule(lr){10-11}
    & & & & & & & & & \textbf{IB} & \textbf{L3} \\
    \midrule
    Ground Truth & / & / & / & / & / & / & / & 0.22 & 0.98 \\
    \midrule
    Cap. (aud)~\cite{mei2023wavcaps} & 2.34 & 0.91 & 1.85 & 14.30 & 9.73 & 1.09 & 0.29 & 1.49 & 0.08 & 0.80 \\
    Cap. (sfx)~\cite{mei2023wavcaps} & 2.09 & 0.86 & 2.05 & 12.53 & 9.12 & 1.14 & 0.16 & 1.51 & 0.09 & 0.81 \\
    Cap. (img)~\cite{li2022blip} & 2.23 & 0.89 & 2.02 & 17.27 & 9.24 & 1.30 & 0.17 & 1.46 & 0.08 & 0.79 \\
    Aud Anlg.~\cite{liu2023audioldm} & 1.37 & 0.77 & 2.34 & 9.33 & 3.97 & 0.91 & 0.12 & 1.53 & 0.11 & 0.82 \\
    S \& R~\cite{petermann2022cocktail} & 1.02 & 0.71 & 2.54 & 8.65 & 3.34 & 0.71 & {\bf 0.11} & 1.51 & 0.12 & 0.83 \\
    AViTAR~\cite{chen2022visual} & 0.76 & 0.32 & 2.56 & 7.44 & 3.02 & 0.68 & 0.17 & 1.47 & 0.14 & 0.87 \\
    \midrule
    Ours & {\bf 0.54} & {\bf 0.20} & {\bf 2.83} & {\bf 5.13} & {\bf 1.64} & {\bf 0.59} & {\bf 0.11} & {\bf 2.03} & {\bf 0.17} & {\bf 0.92} \\
    \bottomrule
    \end{tabular}
    }
    \caption{Quantitative objective results on the \textit{CityWalk} dataset. Captioning (Cap.) can be driven by original conditional audio (aud), separated sound effects (sfx), and conditional images (img). Aud Anlg. and S \& R refer to Audio Analogy and Separate \& Remix respectively. \boldsymbol{\textsuperscript{*}} indicates metrics are evaluated using test set with ground truth.}
    \label{tb:all-obj_result}
    \vspace{-7mm}
\end{table}

\mypar{Baselines.}
We consider a variety of baselines for comparison:
\begin{itemize}[topsep=0pt, noitemsep, leftmargin=*]
    \item {\bf AViTAR} \cite{chen2022visual}: AViTAR is a GAN-based method that is initially proposed to generate indoor room impulse responses conditioned on images, which is not directly compatible with our setting. To address this issue, we integrate an additional audio conditioning branch into this model, and retrain it on our dataset for fair comparison.
    \item {\bf Captioning}: This cascaded approach employs pre-trained image \cite{li2022blip} or audio \cite{mei2023wavcaps} captioning models to generate captions from conditional examples, which are then used to generate sound effects using a text-to-audio model \cite{liu2023audioldm}.
    \item {\bf Audio Analogy} \cite{liu2023audioldm}: AudioLDM is originally used for text-to-audio synthesis. Here we switch the text input with the audio one, allowing us to perform audio-to-audio analogies. We train an audio-visual conditioning model, where we condition on the isolated sound effects instead of the original audio to enforce that the resulting audio does not include any additional speech.
    \item {\bf Separate \& Remix} \cite{petermann2022cocktail}: This method adopts a simple ``copy and paste'' strategy. It uses a pre-trained source separation model to isolate sound effects from the conditional audio, and overlays them onto the input audio at a constant signal-to-noise ratio (SNR) of 8 (we empirically find this can boost both the metrics and auditory perception).
\end{itemize}

\mysubsec{Comparison to Baselines}
\label{subsec:comp_base}
\paragraph{\bf Quantitative results.}
We start by presenting the quantitative results on the {\em CityWalk} dataset in Table~\ref{tb:all-obj_result}. Our model, whether operating in an uni-modal (Table~\ref{tb:comp_uni}) or audio-visual (Table~\ref{tb:all-obj_result}) conditioning, consistently outperforms all the baselines across multiple objective metrics. These results suggest that our model excels in generating more realistic soundscapes compared to the baselines. In particular, Separate \& Remix is worse than our method, despite the fact that it receives nearly the same ambient sounds from the conditional audio. This is probably because our method can not only manipulate ambient sounds but also acoustic properties. We also find that all three Captioning methods perform worse than Separate \& Remix, perhaps due to errors introduced by automatic captioning. Among the Caption-based methods, we observe that using separated sound effects produces more precise captions than the others, leading to the best performance. Although Audio Analogy is trained to resemble the similar ambient sounds of the conditional audio, it still cannot achieve comparable performance to Separate \& Remix, whereas our method can. This highlights the importance of acoustic properties when it comes to soundscape stylization. Moreover, our method surpasses AViTAR, manifesting that the diffusion model excels in producing audio of higher quality compared to the GAN-based counterpart.

\begin{table}[t]
    \scriptsize
    \centering
    \begin{tabular}{l c c c c}
    \toprule
    \textbf{Method} & \textbf{OVL} ($\uparrow$) & \textbf{RAM} ($\uparrow$) & \textbf{RAC} ($\uparrow$) & \textbf{RVI} ($\uparrow$) \\
    \midrule
    Ground Truth & 4.03 $\pm$ 0.09 & / & / & 4.15 $\pm$ 0.11 \\
    \midrule
    Cap. (aud)~\cite{mei2023wavcaps} & 2.58 $\pm$ 0.14 & 2.53 $\pm$ 0.12 & 3.08 $\pm$ 0.10 & 3.13 $\pm$ 0.07 \\
    Cap. (sfx)~\cite{mei2023wavcaps} & 2.77 $\pm$ 0.08 & 3.01 $\pm$ 0.07 & 3.18 $\pm$ 0.13 & 3.22 $\pm$ 0.12 \\
    Cap. (img)~\cite{li2022blip} & 2.14 $\pm$ 0.10 & 2.22 $\pm$ 0.14 & 3.07 $\pm$ 0.15 & 3.09 $\pm$ 0.10 \\
    Aud Analg.~\cite{liu2023audioldm} & 3.08 $\pm$ 0.13 & 3.03 $\pm$ 0.10 & 3.15 $\pm$ 0.11 & 3.12 $\pm$ 0.13 \\
    S \& R~\cite{petermann2022cocktail} & 3.16 $\pm$ 0.09 & 3.34 $\pm$ 0.11 & 3.22 $\pm$ 0.07 & 3.34 $\pm$ 0.08 \\
    AViTAR~\cite{chen2022visual} & 3.32 $\pm$ 0.11 & 3.48 $\pm$ 0.07 & 3.39 $\pm$ 0.12 & 3.40 $\pm$ 0.06 \\
    \midrule
    Ours & {\bf 3.68} $\boldsymbol{\pm}$ {\bf 0.14} & {\bf 3.72} $\boldsymbol{\pm}$ {\bf 0.08} & {\bf 3.55} $\boldsymbol{\pm}$ {\bf 0.09} & {\bf 3.59} $\boldsymbol{\pm}$ {\bf 0.06} \\
    \bottomrule
    \end{tabular}
    \caption{Quantitative subjective results on the \textit{CityWalk} dataset, where OVL, RAM, RAC, and RVI are presented with 95\% confidence intervals. }
    \vspace{-7mm}
    \label{tb:all-sub-results}
\end{table}

To further validate our model's performance, we conduct a human evaluation. We randomly select 100 generated audio samples from the test set, with each sample scored by 40 participants. To prevent random submissions, we include one control set consisting entirely of noise. The participants consistently favor our model's stylized audio, as indicated in Table~\ref{tb:all-sub-results}, which aligns with the objective evaluation results. Interestingly, we observe that the RAC metrics of the first three methods (Captioning, Audio Analogy, and Separate \& Remix) are on par with each other. This consistency could be attributed to the fact that their output is mixed with the same speech as the input, without considering the difference in acoustic properties. This finding also emphasizes the importance of considering acoustic nuances in soundscape stylization, which our model effectively addresses. Furthermore, while AViTAR notably excels beyond other baselines, the inherent challenges in training GAN lead to its audio quality and similarity being consistently worse than those restyled by our method.

\begin{wraptable}[8]{r}{0.585\textwidth} 
\centering
\scriptsize
\vspace{-7.5mm}
\begin{tabular}{lccccc}
\toprule
\multirow{2.5}{*}{\textbf{Method}} & \multicolumn{2}{c}{\bf Seen} & & \multicolumn{2}{c}{\bf Unseen} \\
\cmidrule{2-3} \cmidrule{5-6}
 & {\bf RTE ($\downarrow$)} & {\bf MOSE ($\downarrow$)} & & {\bf RTE ($\downarrow$)} & {\bf MOSE ($\downarrow$)} \\ \midrule
AViTAR~\cite{chen2022visual} & 0.144 & 0.481 & & 0.183 & 0.453 \\ \midrule
Ours & {\bf 0.098} & {\bf 0.412} & & {\bf 0.124} & {\bf 0.399} \\ \bottomrule
\end{tabular}
\vspace{-3.8mm}
\caption{Quantitative comparisons of our method and AViTAR on the {\em Acoustic AVSpeech} dataset.}
\label{tb:quanti-comp}
\end{wraptable}

Additionally, to ensure a fair comparison with AViTAR, we retrain our model using the {\em Acoustic AVSpeech} dataset specifically for the task of visual acoustic matching \cite{chen2022visual}. As illustrated in Table~\ref{tb:quanti-comp}, the quantitative comparison reveals that our method surpasses AViTAR in metrics under seen and unseen settings, thereby demonstrating our method's superior capability in capturing inherent acoustic properties in conditional images.

\begin{figure*}[t]
    \centering
    \includegraphics[width=\linewidth]{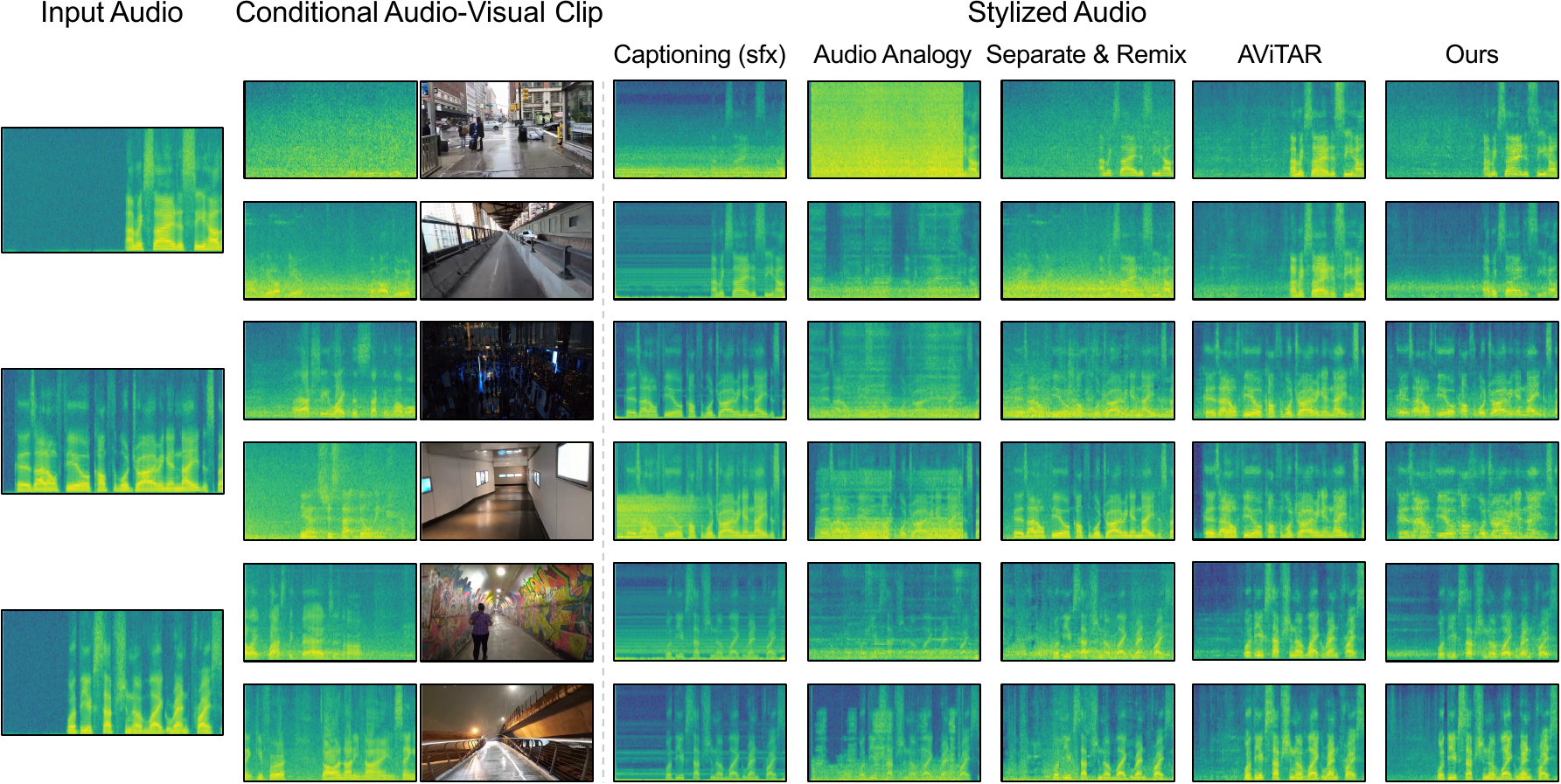}
    \caption{{\bf Model comparison}. We show soundscape stylization results for several models, where each input audio is conditioned on two different audio-visual clips.
    }
    \label{fig:quality}
    \vspace{-6mm}
\end{figure*}

\mypar{Qualitative results.}
We visualize how our results vary under different conditional examples and compare our model with baselines in Figure~\ref{fig:quality}. We also provide additional qualitative results in the Appendix \ref{app:qual}. For the caption-based methods, we only present the best model, which relies on isolated sound effects. Notably, we observe that while these methods occasionally align with the provided conditional examples, they often falter in most instances (like the artifacts introduced in the second example). Audio Analogy appears promising for generating ambient sounds that align with the conditional examples but falls short in complex scenarios, such as the rainy street in the first example. Separate \& Remix tends to directly replicate ambient sounds without considering the specific acoustic environment, leading to less precise acoustic reproduction (like the reverb in the last two examples). AViTAR ranks as the closest approach to our method, but it struggles to capture high frequencies accurately, negatively affecting the overall quality and intelligibility of the output. Our method stands out for its ability to resemble the soundscapes of the conditional example with higher fidelity. For a more direct experience of our model's capabilities, we strongly encourage readers to check out the results video available on the \href{https://tinglok.netlify.app/files/avsoundscape}{project webpage}.

\mysubsec{Ablation Study and Analysis}
Table~\ref{tb:ablations} presents the ablation studies on the {\em CityWalk} dataset. 
We analyze the following model variants: (\romannumeral1) Using the clean audio from the enhancement model as inputs instead of adding Gaussian noise; (\romannumeral2) Employing only the source separation model to isolate the target speech that preserves the original acoustic properties (whereas our default method involves employing a speech enhancement model afterward to modify speech properties); (\romannumeral3) Using only the separated sound effects (no speech) and their corresponding images as conditions; Randomly swapping either (\romannumeral4) the conditional audio, or (\romannumeral5) the conditional image with another at test time to create misaligned audio-visual conditioning; (\romannumeral6) Using only the conditional model (no CFG) to stylize input; (\romannumeral7) Using only the unconditional model to stylize input; (\romannumeral8) Training a ResNet-18-based audio-visual encoder \cite{he2016deep} from scratch rather than using pre-trained CLIP and CLAP. Judging from the results, we draw the following observations:

\mypar{Adding noise mitigates audio nostalgia.}
We ask whether adding noise will help mitigate the effect of ``audio nostalgia'' described in Section~\ref{subsec:diffusion}. Table~\ref{tb:ablations} clearly demonstrates that our approach outperforms the model trained on clean speech by a large margin, providing empirical evidence to support our hypothesis.

\mypar{Acoustic properties play an important role in soundscape stylization.}
We investigate whether our method can resemble plausible acoustic properties to the conditional examples. To examine this, we first train a model with speech extracted from a separation model, matching the acoustic properties of the target and thus excluding the acoustic factor in stylization. Moreover, we train another model conditioned solely on the separated sound effects and their corresponding images. In this scenario, since the conditional audio does not contain speech, it is not sufficiently informative about acoustic variations, leading to arbitrary acoustic changes at test time. As depicted in (\romannumeral2) and (\romannumeral3) of Table~\ref{tb:ablations}, the performances of these variants drastically decline, indicating the critical role of acoustic properties when it comes to soundscape stylization.

\mypar{Visual conditioning complements audio conditioning.}
We explore the impact of visual conditioning on performance. To this end, we create misaligned audio-visual pairs by substituting either the conditional audio or its corresponding image with a random one, and then assess our model's performance with these pairs. As illustrated in (\romannumeral4) and (\romannumeral5) of Table~\ref{tb:ablations}, our model exhibits similar resistance against such perturbations, regardless of whether it is conditioned on the original audio with random images or their inverted versions. This implies that the visual-only model can capture and interpret scene properties, and that visual conditioning delivers complementary information for soundscape stylization than audio conditioning alone.
 
\mypar{Pre-training, CFG, and conditioning enhance stylization.}
We assess the impact of pre-training, CFG, and conditioning on enhancing output relevance. Table~\ref{tb:ablations} demonstrates that our approach outperforms variants that lack these components, demonstrating their effectiveness in improving output relevance.

\begin{table}[t]
    \centering
    \scriptsize
    \begin{tabular}{lcccccc}
    \toprule
     {\bf Method} & {\bf RTE\textsuperscript{*}}($\downarrow$) & {\bf PESQ\textsuperscript{*}}($\uparrow$) & {\bf FD} ($\downarrow$) & {\bf FAD} ($\downarrow$) & {\bf KL} ($\downarrow$) & {\bf IS} ($\uparrow$) \\
    \midrule
     (\romannumeral1) Clean Input & 0.41 & 2.49 & 6.12 & 2.66 & 0.75 & 1.49 \\
     (\romannumeral2) Separation-only Cond. & 0.60 & 2.48 & 6.11 & 2.44 & 0.73 & 1.46 \\
     (\romannumeral3) Sfx-only Cond. & 0.65 & 2.42 & 6.83 & 2.96 & 0.90 & 1.45\\ 
     (\romannumeral4) Random Audio Cond. & 0.71 & 2.20 & 9.51 & 4.92 & 1.05 & 1.44\\ 
     (\romannumeral5) Random Image Cond. & 0.68 & 2.28 & 9.33 & 4.84 & 0.97 & 1.46\\ 
     (\romannumeral6) No CFG & 0.53 & 2.33 & 7.44 & 3.37 & 0.79 & 1.28 \\
     (\romannumeral7) No Cond. & 0.98 & 1.98 & 16.77 & 7.59 & 1.27 & 1.45 \\
     (\romannumeral8) From Scratch & 0.44 & 2.50 & 6.18 & 2.41 & 0.74 & 1.51 \\
    \midrule
     Ours-full & {\bf 0.20} & {\bf 2.83} & \textbf{5.13} & \textbf{1.64} & \textbf{0.59} & \textbf{2.03} \\
    \bottomrule
    \end{tabular}
    \caption{Quantitative ablation studies on the {\em CityWalk} dataset.}
    \vspace{-7mm}
    \label{tb:ablations}
\end{table}

\mysubsec{Cross-Modal and Cross-Domain Evaluation}
\paragraph{\bf Comparison to uni-modal models.}
\label{subsec:comp_unimodal}
We explore the performance of CLAP audio and CLIP image encoders under various conditional settings. Table~\ref{tb:comp_uni} presents a comparison between the fine-tuned audio encoder and its non-fine-tuned counterpart. Notably, fine-tuning significantly enhances performance, suggesting that the original CLAP model has limited generalization capabilities for our dataset. Conversely, fine-tuning the image encoder has little effect on performance gain, indicating its inherent strong generalization abilities. Furthermore, we observe that the fine-tuned audio conditioning model surpasses the image conditioning one, which suggests that audio is a more informative modality for representing soundscapes than visual. Based on these findings, we adopt a late fusion approach \cite{wang2020makes}, combining a fine-tuned CLAP audio encoder with a fixed CLIP image encoder for our audio-visual model. This configuration achieves the best performance among all the models, demonstrating the cross-modal information from the visual modality can help craft more comprehensive soundscapes than the audio modality alone.

\begin{table}[t]
    \centering
    \scriptsize
    \begin{tabular}{cccc|cccccccc}
        \toprule
        {\bf A} & {\bf V} & {\bf FT-A} & {\bf FT-V}  & {\bf RTE}\boldsymbol{\textsuperscript{*}}($\downarrow$) & {\bf PESQ}\boldsymbol{\textsuperscript{*}}($\uparrow$) & {\bf FD} ($\downarrow$) & {\bf FAD} ($\downarrow$) & {\bf KL} ($\downarrow$) & {\bf IS} ($\uparrow$) & {\bf IB} ($\uparrow$) & {\bf L3} ($\uparrow$) \\
        \midrule
        \ding{55} & \ding{51} & \ding{55} & \ding{55} & 0.47 & 2.45 & 6.40 & 2.28 & 0.91 & 1.41 & 0.125 & 0.882 \\
        \ding{55} & \ding{51} & \ding{55} & \ding{51} & 0.44 & 2.39 & 6.39 & 2.29 & 0.91 & 1.40 & 0.123 & 0.884 \\
        \ding{51} & \ding{55} & \ding{55} & \ding{55} & 0.61 & 1.22 & 10.22 & 6.12 & 0.88 & 1.53 & 0.111 & 0.817 \\
        \ding{51} & \ding{55} & \ding{51} & \ding{55} & 0.38 & 2.44 & 5.94 & 2.08 & 0.71 & 1.74 & 0.137 & 0.892 \\
        \ding{51} & \ding{51} & \ding{51} & \ding{55} & \textbf{0.20} & \textbf{2.83} & \textbf{5.13} & \textbf{1.64} & \textbf{0.59} & \textbf{2.03} & \textbf{0.172} & \textbf{0.915} \\          
        \bottomrule
    \end{tabular}
    \caption{Comparison of our uni-modal (A or V) and audio-visual models. A: Audio; V: Visual; FT-A: Fine-tuning for audio; FT-V: Fine-tuning for visual.}
    \vspace{-7mm}
    \label{tb:comp_uni}
\end{table}

\mypar{Generalization to other datasets.}
\begin{wrapfigure}[18]{r}{0.7\textwidth}
    \centering
    \vspace{-8mm}
    \includegraphics[width=\linewidth]{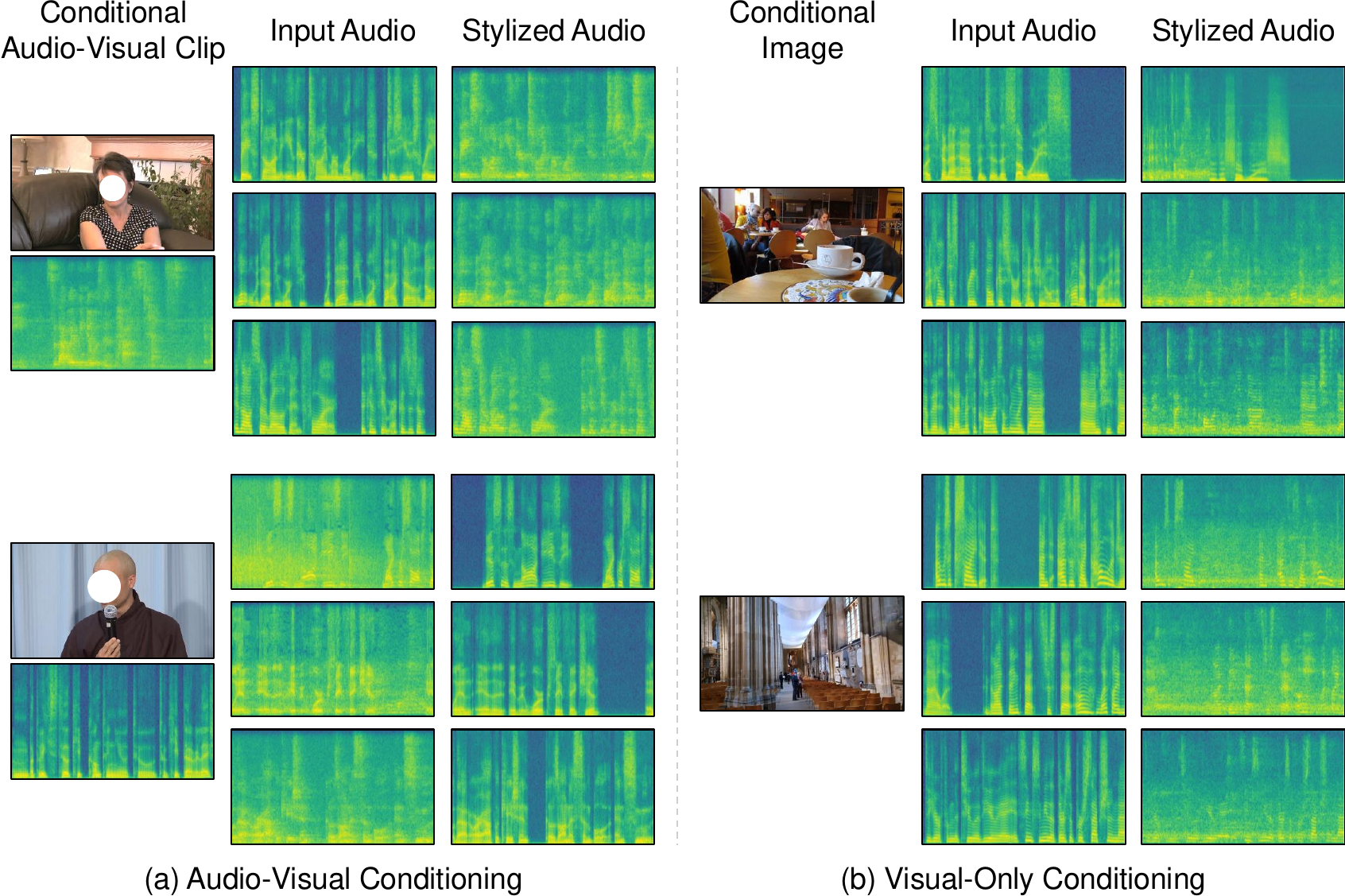}
    \caption{{\bf Qualitative generalization results}. We restyle audio from LRS \cite{son2017lip} conditioned on audio-visual (or visual-only) clips taken from AVSpeech \cite{ephrat2018looking}.}
    \label{fig:generalization}
\end{wrapfigure}

We evaluate the generalization capabilities of our model using out-of-distribution data. Specifically, we explore the model's proficiency in restyling speech from the LRS dataset \cite{son2017lip} conditioned on clips from the AVSpeech dataset \cite{ephrat2018looking}. As illustrated in Figure~\ref{fig:generalization}, we use the clips captured in the indoor room, lecture, coffee shop, and church. We show that our model, whether conditioned on audio-visual or visual-only clips, exhibits robust in-context learning capabilities, effectively adjusting acoustic properties to suit far-field conditions, generating or eliminating ambient sounds, and reducing reverb to enhance speech clarity in alignment with the conditional clips.

\begin{figure}[t]
    \centering
    \begin{minipage}{0.48\textwidth}
        \centering
        \includegraphics[width=\linewidth]{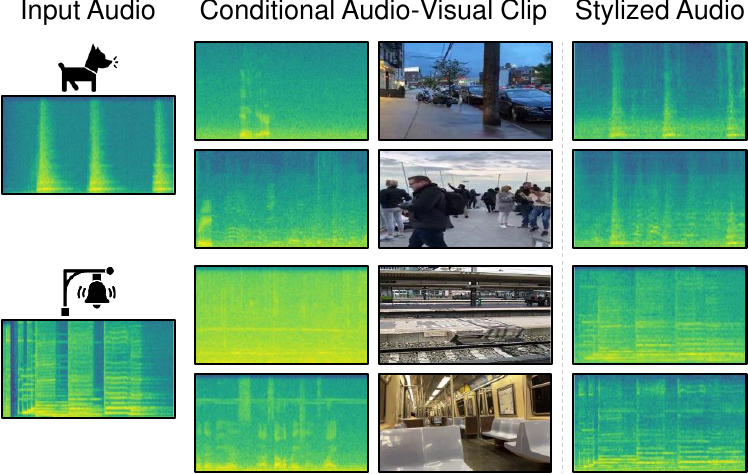}
        \caption{{\bf Generalization to non-speech sounds}. We restyle the sounds of dog barks and train chimes like they are in the conditional scenes.}
        \label{fig:non_speech}
    \end{minipage}\hfill
    \begin{minipage}{0.48\textwidth}
        \centering
        \vspace{-1.5mm}
        \includegraphics[width=\linewidth]{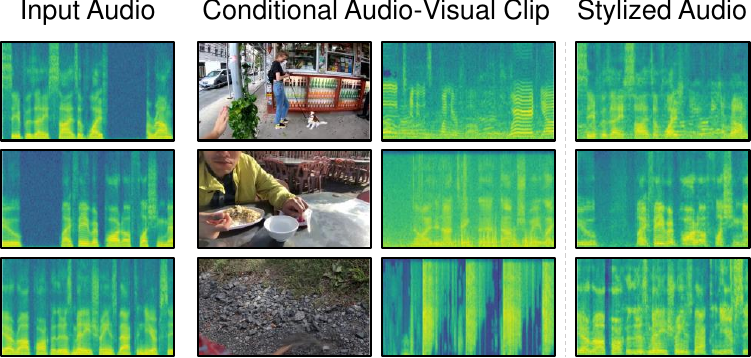}
        \caption{{\bf Failure cases}. Our model fails to resemble the soundscapes of the scenes, perhaps due to vocal effort or invisible sounding objects. It also fails to generate impact sound synchronized with the conditional example.}
        \label{fig:failure}
    \end{minipage}
    \vspace{-4mm}
\end{figure}

\mypar{Generalization to non-speech sounds.}
We examine the adaptability of our model to non-speech sounds. To test this, we introduce sounds such as dog barks and train chimes to the model. As shown in Figure~\ref{fig:non_speech}, we find that our model is capable of stylizing them to emulate soundscapes in streets, viewing platforms, and the interior or exterior of a train, even though it is originally trained solely on speech.

%% file: sec5_discussion.tex
\section{Conclusion}
In this paper, we proposed the {\em audio-visual soundscape stylization} task, aiming to restyle speech to resemble the rich soundscapes from unlabeled in-the-wild audio-visual data. We also constructed an egocentric video dataset and proposed a diffusion model-based method for solving this task in a self-supervised manner. Objective and subjective evaluations demonstrate our model's capability to capture the acoustic properties and ambient sounds of conditional examples. We also show the adaptability of our model to other datasets, visual-only conditioning, and non-speech audio. We hope that our work not only contributes to the task itself but also encourages further exploration into how soundscapes shape our perception of the world. We release the code on our \href{https://tinglok.netlify.app/files/avsoundscape}{project webpage}.

\mypar{Limitations and broader impacts.}
\label{subsec:limitation}

While our model demonstrates promising results across various scenarios, its performance can be inconsistent. As shown in Figure~\ref{fig:failure}, our model struggles with vocal effort challenges \cite{hunter2020toward}, affecting its ability to capture nuances like pitch variations due to speaker-listener distance. Furthermore, when sounding objects are not visually apparent in the conditional clip (e.g., wind sounds), our model may not replicate these sounds accurately. The model also faces difficulties in maintaining audio-visual consistency \cite{chen2021audio}, particularly with nonstationary audio like impact sounds. This highlights the need for model and dataset expansion to enhance scalability. Lastly, while soundscape stylization is useful for content creation such as movie dubbing, it poses a potential risk for creating disinformation videos.

\mypar{Acknowledgements.}
We thank Alexei A. Efros, Justin Salamon, Bryan Russell, Hao-Wen Dong, and Ziyang Chen for their helpful discussions and Baihe Huang for proofreading the paper. This work was funded in part by the Society of Hellman Fellowship and Sony Research Award.

%% file: sec6_appendix.tex
\appendix

\renewcommand{\thesection}{A.\arabic{section}}
\setcounter{section}{0}

\section{Results Video}
\label{app:re_video}
Our results video on the \href{https://tinglok.netlify.app/files/avsoundscape}{project webpage} shows our model's ability to restyle speech to match a variety of input scenes. Additionally, we show:
\begin{itemize}
    \item Despite being trained only on egocentric walking videos, our model can successfully be applied to a variety of out-of-domain speech clips, such as LRS \cite{son2017lip}, AVSpeech \cite{ephrat2018looking}, VGG-Sound \cite{chen2020vggsound}, Into the Wild \cite{li2022learning}, and the classic film {\em Roman Holiday}. 
    \item Our model can generalize to {\em non-speech} sounds taken from BBC Sound Effect \cite{bbc2023sfx}, such as the cry of a baby, a barking dog, train chimes, footsteps, and gunshots. 
    \item We present that the behavior of our model varies with the selected conditional example. 
    \item We find qualitatively that our model can add or remove reverb, transform close-talking and far-field speech, reduce noise, incorporate ambient sounds, and enhance the audio quality of old movies.
\end{itemize}

\section{Dataset Collection}
\label{app:dataset}

\begin{figure}[t]
	\centering
	\begin{subfigure}[h]{0.48\linewidth}
		\centering
		\includegraphics[width=\linewidth]{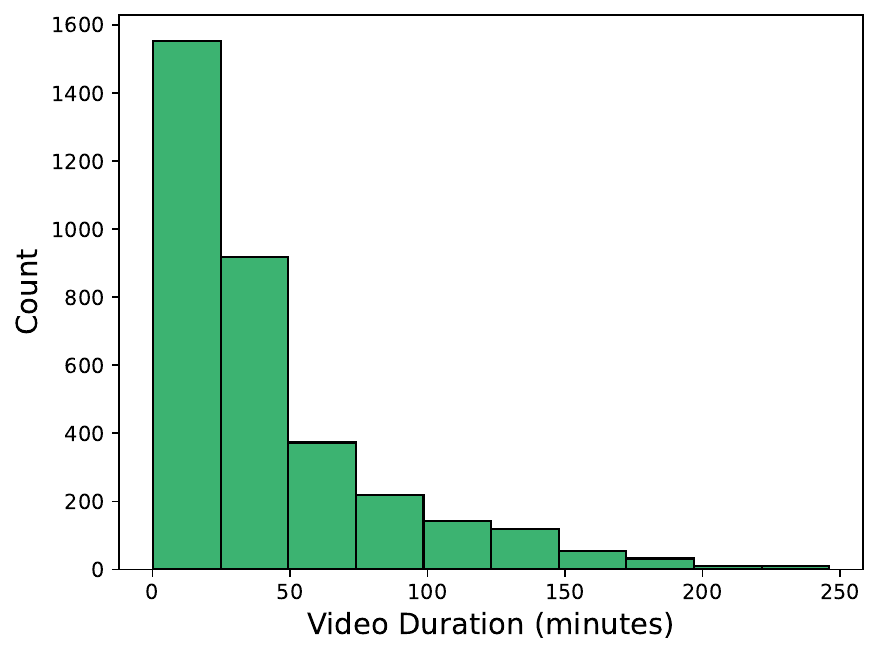}
		\caption{Distribution of video duration}
		\label{fig:duration}
	\end{subfigure}
	\begin{subfigure}[h]{0.48\linewidth}
		\centering
		\includegraphics[width=0.83\linewidth]{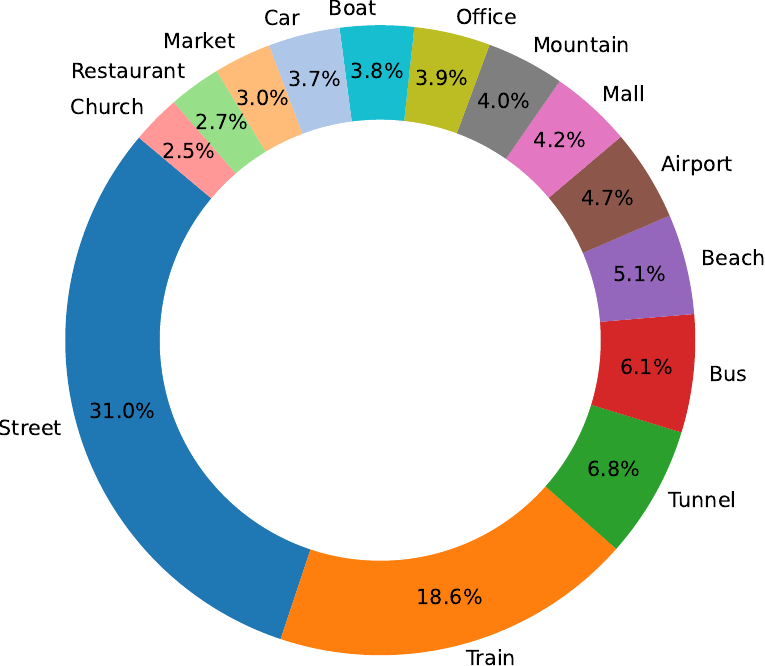}
		\caption{Distribution of video categories}
		\label{fig:categories}
	\end{subfigure}
        \caption{{\bf Statistical analysis of the {\em CityWalk} Dataset}. We present: (a) The distribution of video duration in the datasets; (b) The distribution of the top 14 categories within the dataset deduced by CLIP~\cite{radford2021learning}.}
	\label{fig:dataset}
\end{figure}

\begin{figure}[t]
    \centering
    \includegraphics[width=\linewidth]{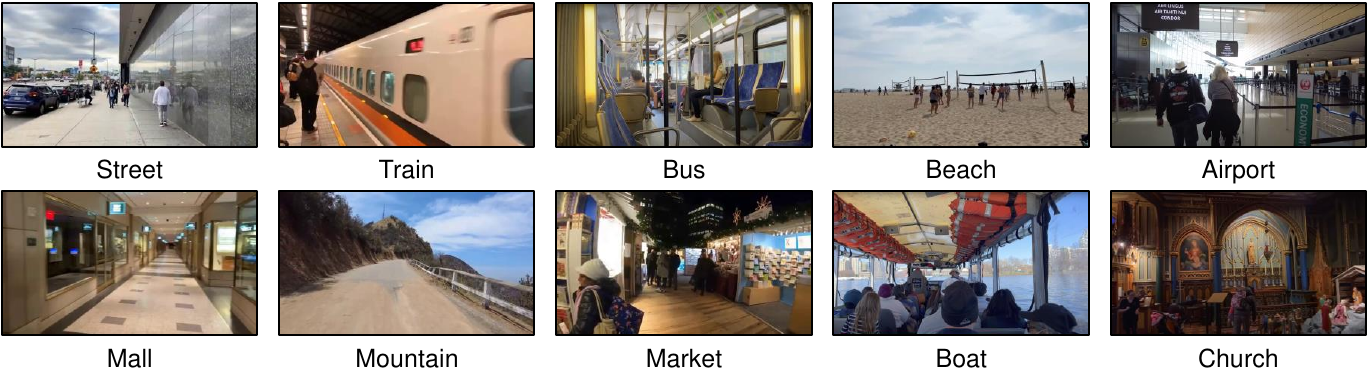}
    \caption{{\bf Example frames of the {\em CityWalk} dataset}. We randomly select 10 different scenes here for showcasing.}
    \label{fig:example_frame}
\end{figure}

We introduce the {\em CityWalk} dataset, a collection of egocentric videos for {\em audio-visual soundscape stylization}. This dataset features a rich diversity of real-world sound textures, which comprises 3,447 indoor (28\%) and outdoor (72\%) videos, with a total length of 2,395 hours. The videos were collected from YouTube, using search queries such as ``City Walk+POV'' and ``City Walk+Binaural.'' Detailed duration statistics are depicted in Figure~\ref{fig:duration}. As illustrated in Figure~\ref{fig:example_frame}, {\em CityWalk} contains a wide spectrum of audio recordings, including human speech and ambient sound, captured in varied environments such as urban streets, train stations, buses, beaches, shopping malls, mountains, markets, and boats, spanning diverse weather conditions. We also provide top-14 categorical distributions in Figure~\ref{fig:categories}, which are acquired from the CLIP \cite{radford2021learning} predictions.

For data filtering and training in our proposed task, we first split each video into 10-second clips and run a pre-trained YAMNet model \cite{plakal2019yamnet} to tag each soundtrack. This step ensures the presence of the targeted audio types within these clips, ensuring that they have not been substituted with alternate sounds, such as voice-overs or background music. Furthermore, we use an off-the-shelf voice activity detector \cite{silerovad2021} to detect speech onsets and exclude silence intervals. The total duration of the {\em CityWalk} dataset is 1,150 hours. We randomly sample 150 hours for model development, allocating 142 hours for training data and reserving the remainder for evaluation without ground truth. Additionally, we sample another 8 hours of held-out videos for assessing metrics between the generated and ground truth audio, achieved by doing the proposed pretext task at test time, \ie, conditioning on different time steps within the same video. Please note that the source of training and testing videos do not overlap.

\begin{figure}[t]
    \centering
    \includegraphics[width=\linewidth]{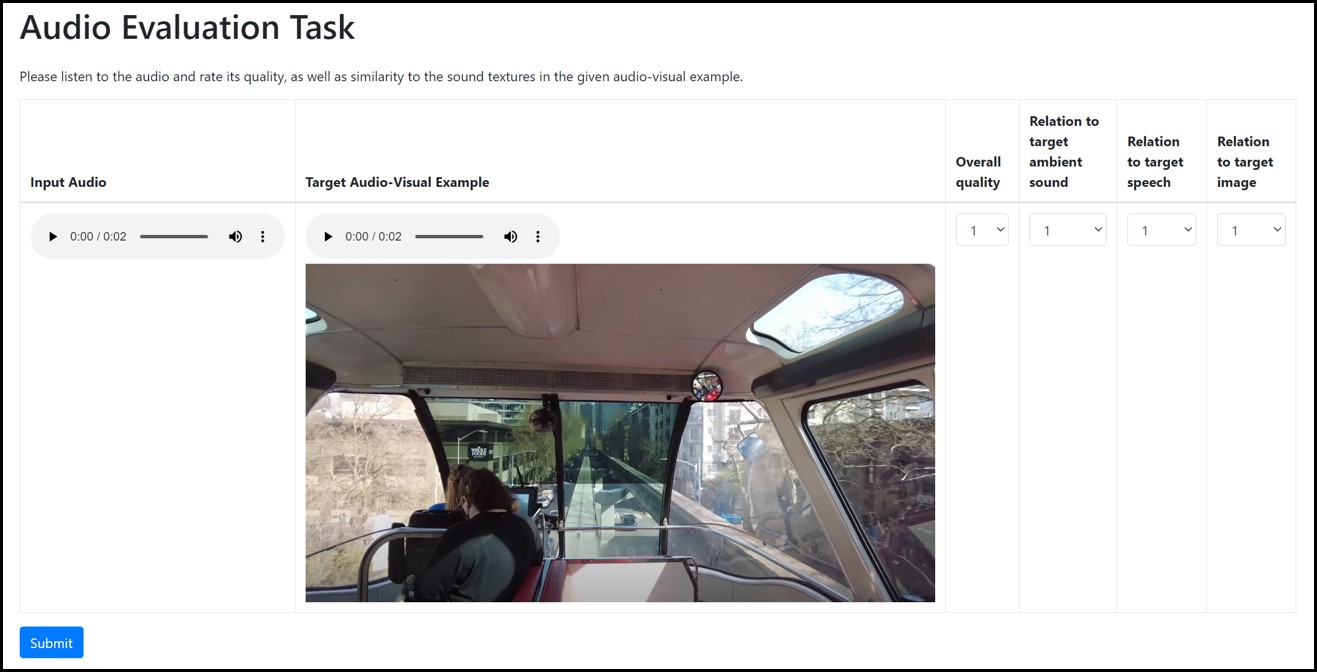}
    \caption{{\bf Interface for Human Evaluation}. We provide a screenshot of the interface designed for evaluating {\em audio-visual soundscape stylization}. Participants are instructed to listen to each audio at least three times, and complete the last four columns prior to advancing to the next example. Upon clicking the ``Submit'' button, participants will be navigated to the next question.}
    \label{fig:mturk}
\end{figure}

\section{Additional Evaluation Details}
\label{app:eval_detail}

\paragraph{\bf RTE.}
We utilize a pre-trained RT60 estimator developed by Chen et al. \cite{chen2022visual}, which processes spectrogram through a ResNet-18 encoder \cite{he2016deep}, to predict the RT60 score. This model is trained on 2.56-second clips of reverberant speech, generated by the SoundSpaces simulator \cite{chen2020soundspaces}, each paired with its corresponding ground truth RT60 value. Training is accomplished by minimizing the MSE loss between the model's predicted RT60 scores and the actual ground truth values. These ground truth RT60 values are determined following the method proposed by Schroeder et al. \cite{schroeder1965new}. For comparison, we report the RT60 difference between each model's output and the ground truth as RTE.

\paragraph{\bf PESQ.}
We evaluate the output speech quality using PESQ. This metric provides an objective measure of speech quality through a comparison of the ground truth and generated speech. The scores range from -0.5 to 4.5, where higher scores indicate better quality. To ensure the reliability of our PESQ evaluations, we perform the tests using the ITU-T P.862 implementation of PESQ. Each input speech is fed to our model and baselines, and the output is then evaluated against the ground truth to compute the score. 

However, it is important to be aware of some drawbacks with the PESQ metric in this context. PESQ is generally designed to take a clean speech reference and measure a degraded speech clip against it. In our case, the reference is a noisy speech signal, and the degraded input is the enhanced noisy speech signal produced by the model. This setup introduces a couple of issues: (\romannumeral1) The phase between the reference and degraded speech will often be inconsistent because HiFi-GAN \cite{kong2020hifi} generates new phases for the vocoded waveforms. And PESQ does have some sensitivity to phase differences. (\romannumeral2) Using PESQ with a noisy reference is likely outside its intended use. This mismatch can affect the accuracy and reliability of the PESQ scores.

\paragraph{\bf Subjective metrics.}
\label{app:sub_eval}
For subjective metrics (OVL, RAM, RAC, RVI) in the main paper, we developed an interface shown in Figure~\ref{fig:mturk}. We selected 100 test samples and each of them was rated by 40 unique participants who are native English speakers to ensure reliability. To maintain anonymity, we organized model outputs in a folder and assigned them with random identifiers. Participants were then tasked with rating each audio file within the context of an audio-visual example by completing the last four columns. We also included a control set containing only white noise to prevent random submissions. Our analysis of the control set revealed consistently low scores given by all human raters, reinforcing the reliability of our evaluation. We also noted that every participant spent a minimum of two minutes on each vote, which strengthened our confidence in the dependability of the results.

\section{Additional Results}
\label{sec: app_add_results}

\paragraph{\bf Adjusting SNR in Separate \& Remix.}
As shown in Table~\ref{tb:app_snr}, we explore the implications of adjusting the signal-to-noise ratio (SNR) in Separate \& Remix baseline \cite{petermann2022cocktail}. Our observations indicate that while tweaking SNR can yield some benefits, the overall impact is relatively limited. It is also important to note this process can be subjective and vary depending on the specific application or user preference. In contrast, our method circumvents the need for such SNR adjustments, thereby demonstrating its robustness and adaptability in various scenarios.

\begin{table}[t]
\scriptsize
\centering
\begin{tabular}{lccccccc}
\toprule
{\bf Method}  & ~{\bf MSE\textsuperscript{*}}({$\downarrow$})~ & ~{\bf FD} ({$\downarrow$})~ & ~{\bf FAD} ({$\downarrow$})~ & ~{\bf KL} ({$\downarrow$})~ & ~{\bf IS} ({$\uparrow$})~ & ~{\bf IB} ({$\uparrow$})~ & ~{\bf L3} ({$\uparrow$}) \\ 
\midrule
SNR=5   & 1.04             & 8.63            & 3.37             & 0.72            & 1.53          & 0.12              & 0.81              \\ 
SNR=8   & 1.02             & 8.65            & 3.34             & 0.71            & 1.51          & 0.12              & 0.83              \\ 
SNR=10  & 1.03             & 8.68            & 3.38             & 0.70            & 1.52          & 0.12              & 0.84              \\ 
\midrule
Origin  & {\bf 1.02}             & 8.87            & 3.48             & 0.70            & 1.51          & 0.11              & 0.82              \\ 
\bottomrule
\end{tabular}
\caption{Quantitative analysis of different SNR levels for Separate \& Remix baseline.}
\label{tb:app_snr}
\end{table}

\paragraph{\bf Enhancement Strategies Comparisons}
\label{app:enhance}
\begin{wrapfigure}[11]{r}{0.6\textwidth}
    \centering
    \vspace{-8.3mm}
    \includegraphics[width=\linewidth]{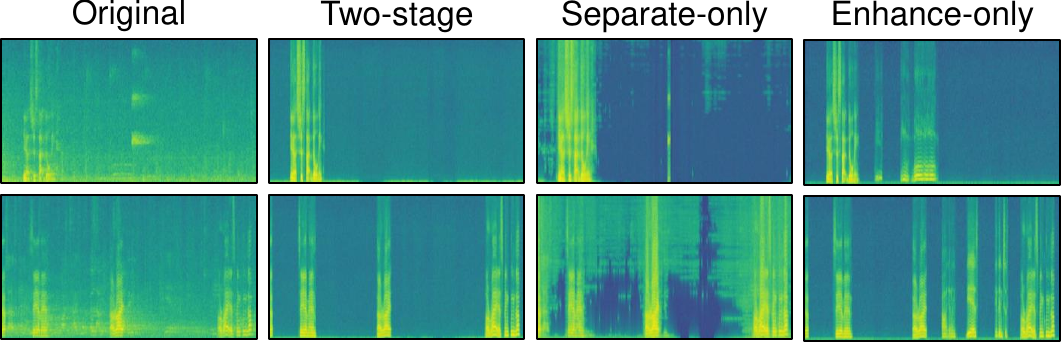}
    \caption{{\bf Qualitative enhancement comparison}. We visualize enhanced audio from different enhancement strategies, including separation-only, enhancement-only, and their combination.}
    \label{fig:separation}
\end{wrapfigure}

We propose a two-stage approach for speech enhancement, consisting of source separation \cite{petermann2022cocktail} as the initial step, followed by speech enhancement \cite{adobe2023enhance} as the subsequent step, to yield the final input audio. This approach is important because relying solely on either source separation or speech enhancement fails to yield the desired speech quality we require.

\begin{wraptable}[7]{r}{0.41\textwidth}
\centering
\scriptsize
\vspace{-8mm}
\begin{tabular}{lcc}
\toprule
\textbf{Method} & \textbf{RT60} ($\downarrow$) & \textbf{OVL} ($\uparrow$)\\
\midrule
Origin & 0.642 & 2.758 \\
Separation-only & 0.487 & 2.974 \\
Enhancement-only & 0.092 & 3.221 \\
Two-stage & \textbf{0.004} & \textbf{3.987} \\
\bottomrule
\end{tabular}
\vspace{-3.5mm}
\caption{Quantitative comparison of different enhancement strategies.}
\label{tb:app_enhance_comp}
\end{wraptable}

As depicted in the penultimate column of Figure~\ref{fig:separation}, using the separation model alone results in isolated speech that retains its original acoustic properties, making our stylization model struggle to acquire the necessary acoustic properties during the training process. Besides, using the enhancement-only method (as seen in the last column of Figure~\ref{fig:separation}) often treats both close-talking and far-field speech as the enhanced target, leading to imprecise foreground speech identification. This, in turn, can degrade the quality of stylization. In contrast, our proposed two-stage approach excels in balancing ambient sounds and acoustic properties (as shown in the second column in Figure~\ref{fig:separation}). We also conduct a quantitative comparison of different enhancement strategies to demonstrate the superiority of our two-stage method in Table~\ref{tb:app_enhance_comp}.

\begin{table}[t]
    \centering
    \scriptsize
    \begin{tabular}{lcccccc}
    \toprule
     {\bf Method} & ~{\bf RTE}\textsuperscript{*}($\downarrow$)~ & ~{\bf PESQ}\textsuperscript{*}($\uparrow$)~ & ~{\bf FD} ($\downarrow$)~ & ~{\bf FAD} ($\downarrow$)~ & ~{\bf KL} ($\downarrow$)~ & ~{\bf IS} ($\uparrow$) \\
    \midrule
     (\romannumeral1) Loudness-norm Cond. & 0.33 & 2.46 & 6.21 & 2.54 & 0.78 & 1.46 \\
     (\romannumeral2) Speech-only Cond. & 0.84 & 2.03 & 13.19 & 6.36 & 1.17 & 1.54 \\
     (\romannumeral3) ISTFT & 0.38 & 2.39 & 6.96 & 3.01 & 0.87 & 1.45 \\
    \midrule
     Ours-full & {\bf 0.20} & {\bf 2.83} & \textbf{5.13} & \textbf{1.64} & \textbf{0.59} & \textbf{2.03} \\
    \bottomrule
    \end{tabular}
    \caption{Additional ablation studies on the {\em CityWalk} dataset.}
    \label{tb:app_ablations}
\end{table}

\paragraph{\bf Additional non-speech sound stylization.}
\begin{wrapfigure}[17]{r}{0.55\textwidth}
    \centering
    \vspace{-8.3mm}
    \includegraphics[width=\linewidth]{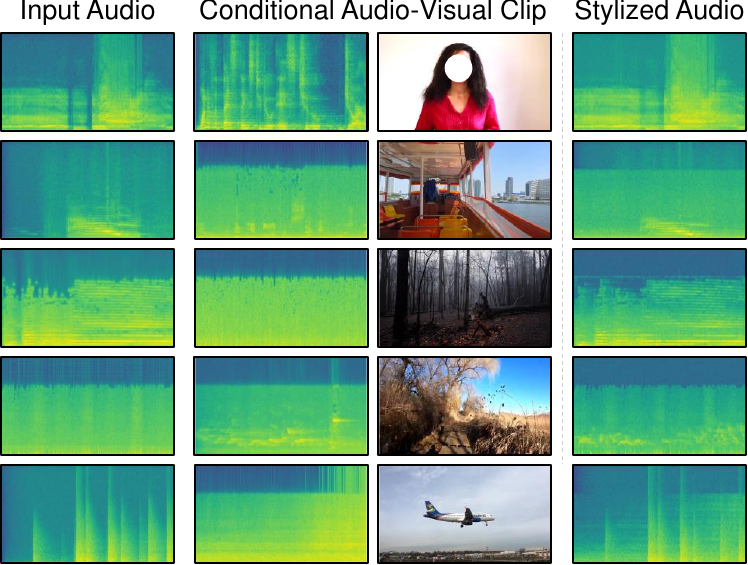}
    \caption{{\bf Additional results on non-speech sound stylization}. We restyle sounds like baby cries and cat mews to match the soundscapes of the specified scenes.}
    \label{fig:app_non_speech}
\end{wrapfigure}

We present additional evidence of our model's adaptability to handle non-speech sounds in Figure~\ref{fig:app_non_speech}. We specifically assess its performance on a variety of sounds, including baby cries, cat meows, footsteps, gunshots, and chicken crows (please note that these sounds are prominent in the foreground). It turns out that our model successfully stylizes these sounds to align with conditional scenes by modifying the room impulse response and generating analogous ambient sounds, including splashing, rain, and the roar of a jet engine, even though it is trained solely on speech data. Please refer to the project webpage for a direct demonstration.

\begin{table}[t]
\scriptsize
\centering
\begin{tabular}{lcccccc}
\toprule
{\bf Method} & {\bf FD} ($\downarrow$) & {\bf FAD} ($\downarrow$) & {\bf KL} ($\downarrow$) & {\bf IS} ($\uparrow$) & {\bf IB} ($\uparrow$) & {\bf L3} ($\uparrow$)\\ \midrule
2.56s A & 5.94 & 2.08 & 0.71 & 1.74 & 0.137 & 0.892 \\ 
5.12s A & 5.89 & 2.10 & 0.72 & 1.75 & 0.137 & 0.894 \\ \midrule
2.56s A + V (Ours) & {\bf 5.13} & {\bf 1.64} & {\bf 0.59} & {\bf 2.03} & {\bf 0.172} & {\bf 0.915} \\ \bottomrule
\end{tabular}
\caption{Quantitative results under different lengths of audio conditioning.}
\vspace{-3mm}
\label{tb:app_audio_context}
\end{table}

\begin{table}[t]
\scriptsize
\centering
\begin{tabular}{lcccccc}
\toprule
{\bf Scale} & ~{\bf FD} ($\downarrow$)~ & ~{\bf FAD} ($\downarrow$)~ & ~{\bf KL} ($\downarrow$)~  & ~{\bf IS} ($\uparrow$)~ & ~{\bf IB} ($\uparrow$)~ & ~{\bf L3} ($\uparrow$)\\
\midrule
$\lambda=1.0$ & 16.77 & 7.59 & 1.27 & 1.45 & 0.082 & 0.694 \\
$\lambda=2.5$ & 6.32 & 2.21 & 0.78 & 1.44 & 0.131 & 0.893 \\
$\lambda=3.5$ & 5.35 & 2.01 & 0.66 & 1.89 & 0.149 & 0.901 \\
$\lambda=4.5$ & {\bf 5.13} & {\bf 1.64} & {\bf 0.59} & {\bf 2.03} & {\bf 0.172} & {\bf 0.915} \\
$\lambda=5.5$ & 5.40 & 1.71 & 0.66 & 1.90 & 0.145 & 0.894 \\
$\lambda=6.5$ & 7.14 & 2.86 & 0.89 & 1.32 & 0.114 & 0.856 \\
\bottomrule
\end{tabular}
\caption{Quantitative results under different CFG scales.}
\label{tb:app_cfg_scale}
\end{table}

\begin{table}[t]
\scriptsize
\centering
\begin{tabular}{l|cccc|c}
\toprule
{\bf Method} & Cap. & Aud Anlg. & S \& R & AViTAR & Ours \\ \midrule
{\bf MS-SNR ($\uparrow$)} & 1.91 & 4.71 & 6.96 & 7.64 & {\bf 8.82} \\ \midrule
\end{tabular}
\vspace{-1mm}
\caption{Magnitude spectrogram SNR results (in dB) of our method and baselines.}
\label{tb:app_ms_snr}
\end{table}

\paragraph{\bf Additional ablation study.}
In Table~\ref{tb:app_ablations}, we introduce two more variants of our method for further analysis: \romannumeral1) Normalizing the loudness of all audio inputs to -20 dB LUFS \cite{steinmetz2021pyloudnorm}; \romannumeral2) Using only the separated speech and its corresponding image frame as conditional examples; \romannumeral3) Employing ISTFT to reconstruct waveform instead of the HiFi-GAN vocoder.

We show that normalizing loudness adversely affects performance. This outcome aligns with our conjecture, given our full model is sensitive to loudness variations. Specifically, our approach is designed to mimic the loudness of the conditional example, which is facilitated by training on audio samples with diverse loudness levels. Imposing a uniform loudness setting restricts the model's ability to adapt to this aspect, thereby reducing its performance. Furthermore, conditioning the model solely on speech, as processed by our enhancement strategy, restricts it to learning only the acoustic properties of speech (no ambient sounds). This limitation hampers the model's performance, as reflected in the notable decline in the quantitative metrics. Furthermore, our findings reveal that employing ISTFT to combine the phase of the input audio for waveform reconstruction detracts from the model's performance, suggesting that the neural vocoder can result in better audio quality.

\paragraph{\bf Different lengths of audio conditioning.}
We test the model with different lengths of audio conditioning. As shown in Table~\ref{tb:app_audio_context}, we find that our model shows no significant improvement under this setting. In contrast, our audio-visual model adds only 1.4\% extra parameters, significantly enhancing performance. This result further suggests that integrating the visual modality can effectively complement the audio modality for representing soundscapes.

\paragraph{\bf Different CFG scales comparisons}
We analyze the performance under various CFG scales ranging from 1 to 6.5. As illustrated in Table~\ref{tb:app_cfg_scale}, we found a steady gain in metrics with $\lambda$ increasing from 1 to 5.5, peaking at 4.5, but declining after then.

\paragraph{\bf Magnitude spectrogram SNR comparisons}
We compare our method with baselines on the magnitude spectrogram SNR (MS-SNR) metric in Table~\ref{tb:app_ms_snr}. Our method outperforms the other approaches.

\begin{figure}[t]
    \centering
    \includegraphics[width=\linewidth]{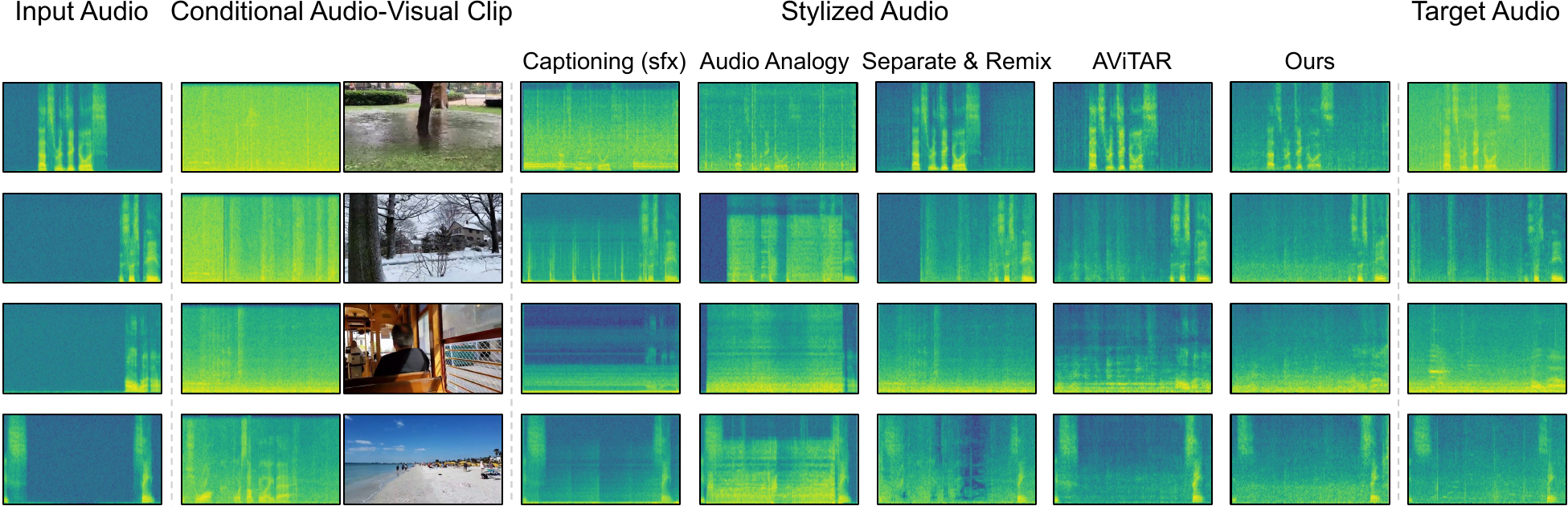}
    \caption{{\bf Additional qualitative results}. We present conditional examples derived from the same video as the input audio, but at different time steps. This is an extension of Figure 4 in the main paper.}
    \vspace{-3.8mm}
    \label{fig:app_qual}
\end{figure}

\paragraph{\bf Additional qualitative comparisons.}
\label{app:qual}
In Figure~\ref{fig:app_qual}, we present additional qualitative comparisons between our approach and the baselines. To provide a comprehensive evaluation, we employ the same held-out video clips at different time intervals as conditional examples, allowing us to illustrate our model's proficiency in reproducing the desired target audio. Furthermore, we introduce conditional examples devoid of speech to facilitate a more precise evaluation of our method and Separate \& Remix.

Specifically, when confronted with non-speech conditional clips, Separate \& Remix \cite{petermann2022cocktail} manages to extract the ambient sounds from the conditioning. However, it struggles to strike an appropriate balance in volume, resulting in the ambient sounds overwhelming the speech, as exemplified in the third case.

Captioning-based methods \cite{mei2023wavcaps, li2022blip} also face challenges when presented with conditional examples devoid of speech. Even in such instances, the generated captions often fall short of capturing the nuanced details within the input. For instance, in the second conditional example, where the conditional audio features footsteps on snow, the generated caption only identifies the presence of footsteps without acknowledging the snow. Consequently, the resulting sound effects deviate from the original conditional example.

Although Audio Analogy \cite{liu2023audioldm} can replicate ambient sounds to some extent, its quality is not as consistent as our approach, probably due to its heavy reliance on isolated ambient sound sources. Furthermore, we find that Audio Analogy occasionally produces large artifacts, as evident in the last three examples.

AViTAR \cite{chen2022visual}, the best-performing prior work, fails to capture the high frequencies of the conditional audio, which leads to a relatively low SNR. One reason for this may be the use of a GAN-based architecture, whereas our approach is based on latent diffusion.

We note that while our method generally outperforms these baselines by considering both ambient sounds and acoustic properties, there are cases where it appears to prioritize the input audio over the conditional one when generating ambient sounds. This may lead to the intensity of the generated soundscape not matching that of the conditional examples. For example, in the first example of Figure~\ref{fig:app_qual}, where the conditioning features heavy rain, our model stylizes the audio to resemble light rain instead, possibly influenced by the mild tone (whisper) of the input audio. This suggests that our model may sometimes place more emphasis on the acoustic properties during the stylization process, resulting in such deviations.